%% file: emnlp2021.tex
\pdfoutput=1

\documentclass[11pt]{article}

\usepackage[]{emnlp2021}
\newcommand{\repo}{\url{https://github.com/robertcsordas/transformer_generalization}}

\usepackage{times}
\usepackage{latexsym}

\usepackage[T1]{fontenc}

\usepackage[utf8]{inputenc}

\usepackage{microtype}
\usepackage{multicol}
\usepackage{multirow}
\usepackage{graphicx}
\usepackage{booktabs, amsmath, amssymb}
\usepackage{xspace}
\usepackage{subfig}
\input{math_commands.tex}

\newcommand{\fig}{Figure}
\newcommand{\tab}{Table}
\newcommand{\sect}{Section}
\newcommand{\teu}{Token Embedding Upscaling}
\newcommand{\ped}{Position Embedding Downscaling}

\newcommand{\EOSp}{\texttt{+EOS}\xspace}
\newcommand{\EOSpOracle}{\texttt{+EOS+Oracle}\xspace}

\newcommand{\EOSmOracle}{\texttt{-EOS+Oracle}\xspace}

\definecolor{mpl_blue}{rgb}{0.122,0.467,0.706}
\definecolor{mpl_green}{rgb}{0.173,0.627,0.173}
\definecolor{mpl_orange}{rgb}{1.0,0.498,0.055}
\definecolor{mpl_red}{rgb}{0.893,0.153,0.157}

\newsavebox\CBox

\title{The Devil is in the Detail: Simple Tricks Improve \\ Systematic Generalization of Transformers}

\author{R\'obert Csord\'as ~ Kazuki Irie ~ J\"urgen Schmidhuber\\
  The Swiss AI Lab IDSIA, USI \& SUPSI, Lugano, Switzerland \\
  \texttt{\{robert, kazuki, juergen\}@idsia.ch}
}

\begin{document}
\maketitle
\begin{abstract}

Recently, many datasets have been proposed to test the
systematic generalization ability of neural networks.
The companion baseline Transformers, typically trained with default hyper-parameters
from standard tasks, are shown to fail dramatically.
Here we demonstrate that by revisiting model configurations
as basic as scaling of embeddings, early stopping, relative positional embedding, and Universal Transformer variants,
we can drastically improve the performance of Transformers on systematic generalization.
We report improvements on five popular datasets: SCAN, CFQ, PCFG, COGS, and Mathematics dataset.
Our models improve accuracy from 50\% to 85\%
on the PCFG productivity split, and
from 35\% to 81\% on COGS.
On SCAN, relative positional embedding largely mitigates the EOS decision problem \citep{newman2020eos}, yielding 100\% accuracy on the length split with a cutoff at 26.
Importantly, performance differences between these models are typically invisible on the IID data split.
This calls for proper generalization validation sets for developing neural networks that generalize systematically.
We publicly release the code to reproduce our results\footnote{\repo}.

\end{abstract}

\section{Introduction}

Systematic generalization \citep{fodor1988connectionism} is a desired property for neural networks
to extrapolate compositional rules seen during training beyond training distribution:
for example, performing different combinations of known rules or 
applying them to longer problems.
Despite the progress of artificial neural networks in recent years, the problem of systematic generalization still remains unsolved \citep{fodor1990connectionism, lake2017generalization, liska2018memorize, greff2020binding, hupkes2019compositionality}.
While there has been much progress in the past years \citep{bahdanau2019closure, korrel2019transcoding, lake2019compositional, li2019compositional, russin2019compositional}, in particular on the popular SCAN dataset \citep{lake2017generalization}
where some methods even achieve 100\% accuracy by introducing some non-trivial symbolic components
into the system \citep{chen2020compositional, liu2020compositional},
the flexibility of such solutions is questionable.
In fact, the existing SCAN-inspired solutions have limited performance gains on other datasets \cite{furrer2020compositional, shaw2020compositional}.
It is thus not enough to solely focus on the SCAN dataset to progress research
on systematic generalization.

Recently, many datasets have been proposed for testing systematic generalization, including PCFG \citep{hupkes2019compositionality} and COGS \citep{kim2020cogs}.
The baseline Transformer models which are released together with the
dataset are typically shown to dramatically fail at the task.
However, the configurations of these baseline models are questionable.
In most cases, some standard practices from machine translation are applied without modification.
Also, some existing techniques such as relative positional embedding \cite{ShawUV18, dai2019transformer}, which are relevant
for the problem, are not part of the baseline.

In order to develop and evaluate methods to improve systematic generalization,
it is necessary to have not only good datasets but also strong baselines to 
correctly evaluate the limits of existing architectures and to avoid false sense of progress over bad baselines.
In this work, we demonstrate that the capability of Transformers \citep{vaswani2017attention} and in particular its universal variants \citep{dehghani2019universal} on these tasks are largely underestimated.
We show that careful designs of model and training configurations are particularly important
for these reasoning tasks testing systematic generalization.
By revisiting configurations such as basic scaling of word and positional embeddings,
early stopping strategy, and relative positional embedding,
we dramatically improve the performance of the baseline Transformers.
We conduct experiments on five datasets: SCAN \citep{lake2017generalization}, CFQ \citep{keysers2020measuring}, PCFG \citep{hupkes2019compositionality}, COGS \citep{kim2020cogs}, and Mathematic dataset \citep{saxton2018analysing}.
In particular, our new models improve the accuracy on the PCFG productivity split from 50\% to 85\%, 
on the systematicity split from 72\% to 96\%,
and on COGS from 35\% to 81\% over the existing baselines.
On the SCAN dataset, we show that our models with relative positional embedding largely mitigates the so-called end-of-sentence (EOS) decision problem \citep{newman2020eos}, achieving 100\% accuracy on the length split with a cutoff at 26.

Also importantly, we show that despite these dramatic performance gaps,
all these models perform equally well on IID validation datasets.
The consequence of this observation is the need for proper generalization validation sets for developing neural networks
for systematic generalization.

We thoroughly discuss guidelines that empirically yield good performance across various datasets,
and we will publicly release the code to make our results reproducible.

\section{Datasets and Model Architectures for Systematic Generalization}

Here we describe the five datasets, and specify the Transformer model variants we use in our experiments.
The selected datasets include both already popular ones and recently proposed ones.
Statistics of the datasets can be found in {\tab} \ref{tab:dataset_stat} in the appendix.\looseness=-1

\subsection{Datasets}

Many datasets in the language domain have been proposed to test systematic generalization.
All datasets we consider here can be formulated as a sequence-to-sequence
mapping task \citep{NIPS2014_5346, graves2012sequence}.
Common to all these datasets, the test set is sampled from a distribution
which is systematically different from the one for training: 
for example, the test set might systematically contain longer sequences,
new combinations or deeper compositions of known rules.
We call this split the \emph{generalization split}. 
Most of the datasets also come with a conventional split, where the train and test (and validation, if available) sets are 
independently and identically distributed samples.
We call this the \emph{IID split}.
In this paper, we consider the following five datasets:

\paragraph{SCAN \cite{lake2017generalization}.}
The task consists of mapping a sentence in natural language into a sequence of commands simulating navigation 
in a grid world.
The commands are compositional: e.g.~an input \texttt{jump twice} should be translated to \texttt{JUMP JUMP}. It comes with multiple data splits:
in addition to the ``simple'' IID split, in the  ``length'' split, the training sequences are shorter than test ones, and in the ``add primitive'' splits,
some commands are presented in the training set only in isolation, without being composed with others. The test set focuses on these excluded combinations.

\paragraph{CFQ \cite{keysers2020measuring}.}
The task consists of translating a natural language question to a Freebase SPARQL query.
For example \texttt{Was M0 a director and producer of M1} should be translated to \texttt{SELECT count(*) WHERE \{M0 ns:film.director.film M1 . M0 ns:film.producer.film | ns:film.production\_company.films M1\}}.
The authors introduce splits based on
``compound divergence'' which measures the difference between the parse trees in the different data splits. The authors experimentally show that it is well correlated with generalization difficulty.
It also comes with a length-based split. \looseness=-1

\paragraph{PCFG \cite{hupkes2019compositionality}.} The task consists of list manipulations and operations that should be executed. For example, \texttt{reverse copy O14 O4 C12 J14 W3} should be translated to \texttt{W3 J14 C12 O4 O14}.
It comes with different splits for testing different aspects of generalization. In this work, we focus on the ``productivity'' split, which focuses on generalization to longer sequences, and on the ``systematicity'' split, which is about recombining constituents in novel ways.

\paragraph{COGS \cite{kim2020cogs}.} The task consists of semantic parsing which maps an English sentence to a logical form.
For example, \texttt{The puppy slept.} should be translated to \texttt{* puppy ( x \_ 1 ) ; sleep .~agent ( x \_ 2, x \_ 1 )}.
It comes with a single split, with a training, IID validation and OOD generalization testing set. \looseness=-1

\paragraph{Mathematics Dataset \cite{saxton2018analysing}.} The task consists of high school level textual math questions, e.g. \texttt{What is -5 - 110911?} should be translated to \texttt{-110916}. 
The data is split into different subsets by the problem category, called modules.
Some of them come with an extrapolation set, designed to measure generalization. 
The amount of total data is very large and thus expensive to train on,
but different modules can be studied individually.
We focus on ``add\_or\_sub" and ``place\_value" modules.

\subsection{Model Architectures}\label{sec:model_architectures}
We focus our analysis on two Transformer architectures: standard Transformers \cite{vaswani2017attention} and Universal Transformers \cite{dehghani2019universal},
and in both cases with absolute or relative positional embedding \cite{dai2019transformer}. 
Our Universal Transformer variants are simply Transformers with shared weights between layers, without adaptive computation time \cite{schmidhuber2012self, graves2016adaptive} and timestep embedding.
Positional embedding are only added to the first layer.

Universal Transformers are particularly relevant for reasoning and algorithmic tasks.
For example, if we assume a task which consists in executing a sequence of operations,
a regular Transformer will learn successive operations in successive layers with separate weights.
In consequence, if only some particular orderings of the operations are seen during training,
each layer will only learn a subset of the operations,
and thus, it will be impossible for them to recombine operations in an arbitrary order.
Moreover, if the same operation has to be reused multiple times, the network has to re-learn it, which is harmful for systematic generalization and reduces the data efficiency of the model \cite{csordas2021are}. 
Universal Transformers have the potential to overcome this limitation: sharing the weights between each layer makes it possible to reuse the existing knowledge from different compositions. On the downside, the Universal Transformer's capacity can be limited because of the weight sharing.

\section{Improving Transformers on Systematic Generalization}

In this section, we present methods
which greatly improve Transformers on systematic generalization tasks,
while they could be considered as details in standard tasks.
For each method, we provide experimental evidences on a few representative datasets.
In {\sect}~\ref{sec:results}, we apply these findings to all datasets.

\subsection{Addressing the EOS Decision Problem with Relative Positional Embedding}\label{sec:eos_decision}

\begin{table*}
    \centering
    \small
    \caption{Exact match accuracies on length splits with different cutoffs. Reported results are the median of 5 runs.
    Trafo denotes Transformers.
    The numbers in the rows \EOSpOracle and \EOSmOracle are taken from \citet{newman2020eos} as
    reference numbers but they can not be compared to others as they are evaluated with oracle length.
    Our models use different hyperparameters compared to theirs.
    We refer to {\sect}~\ref{sec:eos_decision} for details.}
    \label{tab:scan-results}
    \begin{tabular}{c l r r r r r r r r r r r}
        \toprule
        & $\ell$ \textit{(length cutoff)} & 22 & 24 & 25 & 26 & 27 & 28 & 30 & 32 & 33 & 36 & 40 \\
        \midrule
        \parbox[t]{3mm}{\multirow{3}{*}{\rotatebox[origin=c]{90}{\tiny Reference}}} & \EOSp & 0.00 & 0.05 & 0.04 & 0.00 & 0.09 & 0.00 & 0.09 & 0.35 & 0.00 & \it0.00 & 0.00\\
        & \it\EOSpOracle & \it0.53 & \it0.51 & \it0.69 & \it0.76 & \it0.74 & \it0.57 & \it0.78 & \it0.66 & \it0.77 & \textit{\textbf{1.00}} & \it0.97\\
         & \it\EOSmOracle & \textit{\textbf{0.58}} & \textit{\textbf{0.54}} & \it0.67 & \it0.82 & \it0.88 & \it0.85 & \it0.89 & \it0.82 & \textit{\textbf{1.00}} & \textit{\textbf{1.00}} & \textit{\textbf{1.00}}\\
         \midrule
         \parbox[t]{3mm}{\multirow{4}{*}{\rotatebox[origin=c]{90}{\tiny Ours (+EOS)}}} &
        Trafo \xspace & 0.00 & 0.04 & 0.19 & 0.29 & 0.30 & 0.08 & 0.24 & 0.36 & 0.00 & 0.00 & 0.00 \\
 & \quad + Relative PE  \xspace & 0.20 & 0.12 & 0.31 & 0.61 & \bf1.00 & \bf1.00 & \bf1.00 & 0.94 & \bf1.00 & \bf1.00 & \bf1.00 \\
 & Universal Trafo \xspace & 0.02 & 0.05 & 0.14 & 0.21 & 0.26 & 0.00 & 0.06 & 0.35 & 0.00 & 0.00 & 0.00 \\
 & \quad + Relative PE \xspace & 0.20 & 0.12 & \bf0.71 & \bf1.00 & \bf1.00 & \bf1.00 & \bf1.00 & \bf1.00 & \bf1.00 & \bf1.00 & \bf1.00 \\       
        \bottomrule
    \end{tabular}
\end{table*}

\paragraph{The EOS decision problem.}
A thorough analysis by \citet{newman2020eos} highlights that LSTMs and Transformers struggle to generalize to longer output lengths than they are trained for. Specifically, it is shown that the decision when to end the sequence (the EOS decision) often overfits to the specific positions observed in the train set.
To measure whether the models are otherwise able to solve the task, they conduct a so-called \textit{oracle evaluation}: they ignore the EOS token during evaluation, and use the ground-truth sequence length to stop decoding. 
The performance with this evaluation mode is much better, which illustrates that the problem is indeed the EOS decision.
More surprisingly, if the model is trained without EOS token as part of output vocabulary
(thus it can only be evaluated in oracle mode), the performance is further improved.
It is concluded that teaching the model when to end the sequence has undesirable side effects on the model's length generalization ability.

We show that the main cause of this EOS decision problem in the case of
Transformers is the absolute positional embedding.
Generally speaking, the meaning of a word is rarely dependent on the word's absolute position in a document but depends on its neighbors.
Motivated by this assumption, various relative positional embedding methods \cite{ShawUV18, dai2019transformer}
have been proposed.
Unfortunately, they have not been
considered
for systematic generalization in prior work
(however, see Sec.~\ref{sec:related}), even though they are particularly relevant for that.

We test Transformers with relative positional embedding in the form used in Transformer XL \cite{dai2019transformer}.
Since it is designed for auto-regressive models,
we directly apply it in the decoder of our model, while for the
encoder, we use a symmetrical variant of it (see Appendix \ref{sec:rel_pos}).
The interface between encoder and decoder uses the standard attention without any positional embedding.

Our experimental setting is similar to \citet{newman2020eos}.
The length split in SCAN dataset restricts the length of the train samples to 22 tokens (the test set consists of samples with an output of more than 22 tokens).
This removes some compositions from the train set entirely, which introduces
additional difficulty to the task.
$80\%$ of the test set consists of these missing compositions.
In order to mitigate the issue of unknown composition and focus purely on the length problem, \citet{newman2020eos} re-split SCAN by introducing different length cutoffs and report the performance of each split. 
We test our models similarly. However, our preliminary experiments showed the performance of the original model is additionally limited by being too shallow: it uses only 2 layers for both the encoder and decoder. We increased the number of layers to 3. To compensate for the increased number of parameters, we decrease the size of the feed-forward layers from 1024 to 256. In total, this reduces the number of parameters by $30\%$. We train our models with Adam optimizer, a learning rate of $10^{-4}$, batch size of 128 for 50k steps.

The results are shown in Table \ref{tab:scan-results}.
In order to show that our changes of hyperparameters are not the main reason for the improved performance, we report the performance of our modified model without relative positional embedding (row Trafo).
We also include the results from \citet{newman2020eos} for reference.
We report the performance of Universal Transformer models trained with identical hyperparameters.
All our models are trained \emph{to predict} the EOS token and are evaluated \emph{without} oracle (\EOSp configuration).
It can be seen that both our standard and Universal Transformers with absolute positional embedding have near-zero accuracy for all length cutoffs, whereas models with relative positional embedding excel: they even outperform the models trained without EOS prediction and evaluated with the ground-truth length.

Although Table \ref{tab:scan-results} highlights the advantages of relative positional embedding and shows that they can largely mitigate the EOS-overfitting issue,
this does not mean that the problem of generalizing to longer sequences is fully solved.
The sub-optimal performance on short length cutoffs (22-25) indicates that the model finds it hard to zero-shot generalize to unseen compositions of specific rules. To improve these results further, research on models which assume analogies between rules and compositions are necessary, such that they can recombine known constituents without any training example. 

\paragraph{Further benefits of relative positional embedding.}\label{sec:further_benefits}
In addition to the benefit highlighted in the previous paragraph,
we found that models with relative positional embedding are easier to train in general.
They converge faster ({\fig} \ref{fig:relative} in the appendix) and are less sensitive to batch size (Table \ref{tab:small_batch_details_abs} in the appendix).
As another empirical finding, we note that relative Transformers without shared layers sometimes catastrophically fail before reaching their final accuracy: the accuracy drops to 0, and it never recovers. We observed this with PCFG productivity split and the ``Math:~place\_value'' task. Reducing the number of parameters (either using Universal Transformers or reducing the state size) usually stabilizes the network. \looseness=-1

\subsection{Model Selection Should Be Done Carefully}

\label{sec:early_stopping}
\begin{figure}
\begin{center}

\includegraphics[width=0.95\columnwidth]{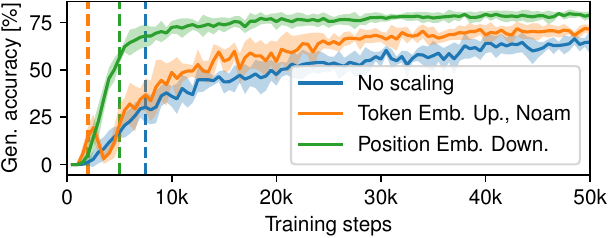}
\qquad
\caption{Generalization accuracy on COGS as a function of training steps for standard Transformers
with different embedding scaling schemes.
The vertical lines show the median of the early stopping points for the five runs.
Early stopping parameters are from \citet{kim2020cogs}.
``Token Emb. Up., Noam'' corresponds to the baseline configuration \cite{kim2020cogs}.
See Sec.~\ref{sec:initialization} for details on scaling.
}
\label{fig:early_stopping}
\end{center}
\end{figure}

\paragraph{The danger of early stopping.}
Another crucial aspect greatly influencing the generalization performance of Transformers
is model selection, in particular early stopping.
In fact, on these datasets, it is a common practice to use only the IID split to tune hyperparameters or select models with early stopping (e.g.~\citet{kim2020cogs}).
However, since any reasonable models achieve nearly 100\% accuracy on the IID validation set,
there is no good reason to believe this to be a good practice for selecting models
for generalization splits.
To test this hypothesis, we train models on COGS dataset
without early stopping, but with a fixed number of 50k training steps.
The best model achieved a test accuracy of $81\%$,
while the original performance by \citet{kim2020cogs} is $35\%$.
Motivated by this huge performance gap, we had no other choice but to conduct an analysis
on the generalization split
to demonstrate the danger of early stopping and
discrepancies between the performance on the IID and generalization split.
The corresponding results are shown in {\fig} \ref{fig:early_stopping}
(further effect of embedding scaling is discussed in next Sec.~\ref{sec:initialization})
and {\tab} \ref{tab:init_iid}.
Following \citet{kim2020cogs}, we measure the model's performance every 500 steps, and mark the point where early stopping with patience of 5 would pick the best performing model. It can be seen that in some cases the model chosen by early stopping is not even reaching half of the final generalization accuracy.

\begin{table}
    \centering
    \small
    \caption{Final IID validation and generalizations accuracy for COGS (50k steps) and PCFG Productivity set (300k steps) with different scaling ({\sect} \ref{sec:initialization}). {\teu} (TEU) is unstable on PCFG with our hyperparameters.
    {\ped} (PED) performs the best on both datasets.
    }
    \label{tab:init_iid}
    \begin{tabular}{l l c c}
        \toprule
 & & IID Validation & Gen. Test\\
\midrule
\parbox[t]{3mm}{\multirow{3}{*}{\rotatebox[origin=c]{90}{\small COGS}}}
 & TEU & {\bf1.00 $\pm$ 0.00} & 0.78 $\pm$ 0.03 \\
 & No scaling & {\bf1.00 $\pm$ 0.00} & 0.62 $\pm$ 0.06 \\
 & PED & {\bf1.00 $\pm$ 0.00} & {\bf0.80 $\pm$ 0.00} \\
\midrule
\parbox[t]{3mm}{\multirow{3}{*}{\rotatebox[origin=c]{90}{\small PCFG}}}
 & TEU & 0.92 $\pm$ 0.07 & 0.47 $\pm$ 0.27 \\
 & No scaling & {\bf0.97 $\pm$ 0.01} & 0.63 $\pm$ 0.02 \\
 & PED & 0.96 $\pm$ 0.01 & {\bf0.65 $\pm$ 0.03} \\
\bottomrule
    \end{tabular}
\end{table}

To confirm this observation in the exact setting of \citet{kim2020cogs},
we also disabled the early stopping in the original codebase
\footnote{\url{https://github.com/najoungkim/COGS}}, and observed that the accuracy improved to $65\%$ without any other tricks.
We discuss further performance improvements on COGS dataset in {\sect} \ref{sec:cogs}. 

\begin{figure}[t]
\begin{center}
\includegraphics[width=0.95\columnwidth]{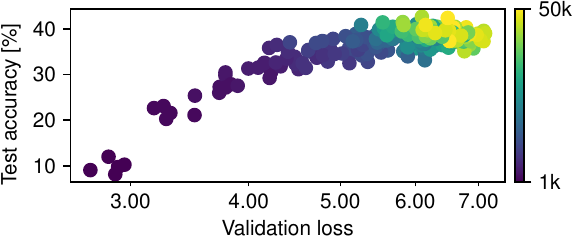}

\caption{Relationship between validation loss and test accuracy (same distribution) on CFQ MCD 1 split
for a relative Transformer. The color shows the training step. Five runs are shown. The loss has a logarithmic scale. High accuracy corresponds to higher loss, which is unexpected. For detailed analysis, see {\fig} \ref{fig:loss_grow_analysis}.
}\label{fig:loss_grows}

\end{center}
\end{figure}

\paragraph{The lack of validation set for the generalization split.}

A general problem raised in the previous paragraph is the lack
of validation set for evaluating models for generalization.
Most of the datasets come without a validation set for the generalization split (SCAN, COGS, and PCFG). Although CFQ comes with such a set, the authors argue that only the IID split should be used for hyperparameter search, and it is not clear what should be used for model development.

In order to test novel ideas, a way to gradually measure progress is necessary, such that the effect of changes can be evaluated. If the test set is used for developing the model, it implicitly risks overfitting to this test set. On the other hand, measuring performance on the IID split does not necessarily provide any valuable information about the generalization performance on the systematically different test set (see Table \ref{tab:init_iid}). The IID accuracy of all the considered datasets is $100\%$ (except on PCFG where it's also almost $100\%$); thus, no further improvement,
nor potential difference between generalization performance of models
can be measured (see also {\tab} \ref{tab:results_iid} in the appendix).

It would be beneficial if future datasets would have a validation and test set for both the IID \emph{and} the generalization split.
For the generalization split, the test set could be designed to be more difficult than the validation set.
This way, the validation set can be used to measure progress during development,
but overfitting to it would prevent the model to generalize well to the test set.
Such a division can be easily done on the splits for testing productivity.
For other types of generalization, we could use multiple datasets
sharing the same generalization problem.
Some of them could be dedicated for development and others for testing.

\paragraph{Intriguing relationship between generalization accuracy and loss.}\label{sec:loss_and_accuracy}
Finally, we also note the importance of using accuracy (instead of loss)
as the model selection criterion.
We find that the generalization accuracy and loss do not necessarily correlate,
while sometimes, model selection based on the loss is reported in practice e.g.~in \citet{kim2020cogs}.
Examples of this undesirable behavior are shown on {\fig} \ref{fig:loss_grows} for CFQ and on {\fig} \ref{fig:loss_grows_details} in the appendix for COGS dataset. On these datasets, the loss and accuracy on the generalization split both grows during training.
We conducted an analysis to understand the cause of this surprising phenomenon,
we find that the total loss grows because the loss of the samples with incorrect outputs increases more than it improves on the correct ones.
For the corresponding experimental results, we refer to {\fig} \ref{fig:loss_grow_analysis} in the appendix.
We conclude that even if a validation set is available for the generalization split,
it would be crucial to use the \emph{accuracy instead of the loss} for early stopping and hyperparameter tuning. \looseness=-1

\begin{figure}
\begin{center}
\includegraphics[width=0.95\columnwidth]{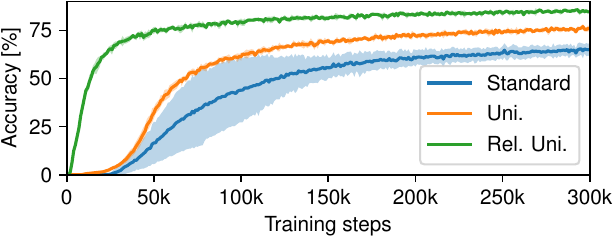}\\
\includegraphics[width=0.95\columnwidth]{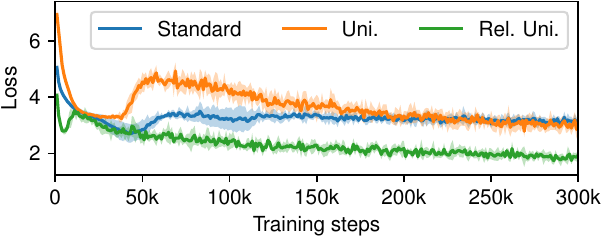}\\

\caption{Test loss and accuracy on PCFG during training. The loss exhibits an epoch-wise double descent phenomenon \cite{nakkiran2019deep}, while the accuracy increases monotonically. {\color{mpl_blue} {Standard Transformer}} with PED (Sec.~\ref{sec:initialization}), {\color{mpl_orange}{Universal Transformer with absolute}}, and {\color{mpl_green} {relative positional embeddings}} are shown.
} \label{fig:pcfg_double_descent}
\end{center}
\end{figure}

Finally, on PCFG dataset, we observed epoch-wise double descent phenomenon \cite{nakkiran2019deep}, as shown in {\fig}  \ref{fig:pcfg_double_descent}.
This can lead to equally problematic results if the loss is used for model selection or tuning.

\subsection{Large Impacts of Embedding Scaling}\label{sec:initialization}

The last surprising detail which greatly influences
generalization performance
of Transformers is the choice of embedding scaling scheme. 
This is especially important for
Transformers with absolute positional embedding, where the word and positional embedding have to be combined.
We experimented with the following scaling schemes:
\begin{enumerate}
    \item {\teu} (TEU). This is the standard scaling used by \citet{vaswani2017attention}. It uses Glorot initialization \cite{glorot2010understanding} for the word embeddings. However, the range of the sinusoidal positional embedding is always in $[-1,1]$. Since the positional embedding is directly added to the word embeddings, this discrepancy can make the model untrainable. Thus, the authors upscale the word embeddings by $\sqrt{d_{\text{model}}}$ where $d_{\text{model}}$ is the embedding size.
OpenNMT\footnote{\url{https://opennmt.net/}}, the framework used for the baseline models for PCFG and COGS datasets respectively by \citet{hupkes2019compositionality} and \citet{kim2020cogs}, also uses this scaling scheme. 
    \item No scaling. It initializes the word embedding with $\mathcal{N}(0,1)$ (normal distribution with mean 0 and standard deviation of 1). Positional embeddings are added without scaling.
    \item {\ped} (PED), which uses Kaiming initialization \cite{he2015delving}, and scales the positional embeddings by $\frac{1}{\sqrt{d_{\text{model}}}}$.
\end{enumerate}

The PED differs from TEU used in \citet{vaswani2017attention} in two ways:
instead of scaling the embedding up,
PED scales the positional embedding down and uses Kaiming instead of Glorot initialization.
The magnitude of the embeddings should not depend on the number of words in the vocabulary but on the embedding dimension.

{\tab} \ref{tab:init_iid} shows the results.
Although ``no scaling'' variant is better than TEU on the PCFG test set, it is worse on the COGS test
set.
PED performs consistently the best on both datasets. 
Importantly, the gap between the best and worst configurations is large
on the test sets.
The
choice of scaling thus also contributes in
the large improvements we report over the existing baselines.\looseness=-1

\section{Results Across Different Datasets}

\label{sec:results}
In this section, we apply the methods we illustrated in the previous section
across different datasets.
{\tab} \ref{tab:results} provides an overview of all improvements we obtain
on all considered datasets.
Unless reported otherwise, all results are the mean and standard deviation of 5 different random seeds.
If multiple embedding scaling schemes are available, we pick the best performing one for a fair comparison. Transformer variants with relative positional embedding outperform the absolute variants on almost all tested datasets. Except for COGS and CFQ MCD 1, the universal variants outperform the standard ones.
In the following, we discuss and highlight the improvements we obtained for each individual dataset.

\begin{table*}
    \centering
    \small
    \caption{Test accuracy of different Transformer (Trafo) variants on the considered datasets. See Sec.~\ref{sec:results} for details. The last column shows previously reported accuracies. References: [1] \citet{newman2020eos}, [2] \citet{keysers2020measuring},  [3] \url{https://github.com/google-research/google-research/tree/master/cfq}, [4] \citet{hupkes2019compositionality}, [5] \citet{kim2020cogs}, [6] \citet{saxton2018analysing}. Results marked with $*$ cannot be directly compared because of different training setups. $\sim$ denotes approximative numbers read from charts reported in previous works.}
    \label{tab:results}
    \begin{tabular}{l c c c c | c}
        \toprule
  & Trafo & Uni. Trafo & Rel. Trafo & Rel. Uni. Trafo & Prior Work \\
\midrule
SCAN (length cutoff=26) & 0.30 $\pm$ 0.02 & 0.21 $\pm$ 0.01 & 0.72 $\pm$ 0.21 & \bf{1.00 $\pm$ 0.00} & $0.00^{[1]}$ \\
\midrule
CFQ Output length & 0.57 $\pm$ 0.00 & 0.77 $\pm$ 0.02 & 0.64 $\pm$ 0.06 & \bf{0.81 $\pm$ 0.01} & $\sim 0.66^{[2]}$ \\
CFQ MCD 1 & \bf{0.40 $\pm$ 0.01} & 0.39 $\pm$ 0.03 & 0.39 $\pm$ 0.01 & 0.39 $\pm$ 0.04 & $0.37\pm0.02^{[3]}$ \\
CFQ MCD 2 & 0.10 $\pm$ 0.01 & 0.09 $\pm$ 0.02 & 0.09 $\pm$ 0.01 & \bf{0.10 $\pm$ 0.02} & $0.08\pm0.02^{[3]}$ \\
CFQ MCD 3 & 0.11 $\pm$ 0.00 & 0.11 $\pm$ 0.01 & 0.11 $\pm$ 0.01 & \bf{0.11 $\pm$ 0.03} & $0.11\pm0.00^{[3]}$ \\
CFQ MCD mean & 0.20 $\pm$ 0.14 & 0.20 $\pm$ 0.14 & 0.20 $\pm$ 0.14 & \bf{0.20 $\pm$ 0.14} & $0.19\pm0.01^{[2]}$ \\
\midrule
PCFG Productivity split & 0.65 $\pm$ 0.03 & 0.78 $\pm$ 0.01 & - & \bf{0.85 $\pm$ 0.01} & $0.50\pm0.02^{[4]}$ \\
PCFG Systematicity split & 0.87 $\pm$ 0.01 & 0.93 $\pm$ 0.01 & 0.89 $\pm$ 0.02 & \bf{0.96 $\pm$ 0.01} & $0.72\pm0.00^{[4]}$ \\
\midrule
COGS & 0.80 $\pm$ 0.00 & 0.78 $\pm$ 0.03 & \bf{0.81 $\pm$ 0.01} & 0.77 $\pm$ 0.01 & $0.35\pm0.06^{[5]}$ \\
\midrule
Math: add\_or\_sub & 0.89 $\pm$ 0.01 & 0.94 $\pm$ 0.01 & 0.91 $\pm$ 0.03 & \bf{0.97 $\pm$ 0.01} & $\sim0.91^{[6]*}$ \\
Math: place\_value & 0.12 $\pm$ 0.07 & 0.20 $\pm$ 0.02 & - & \bf{0.75 $\pm$ 0.10} & $\sim0.69^{[6]*}$ \\
        \bottomrule
    \end{tabular}
\end{table*}

\subsection{SCAN}

We focused on the \textbf{length split} of the dataset.
We show that it is possible to mitigate the effect of overfitting to the absolute position of the EOS token by using relative positional embedding. We already discussed the details in Sec.~\ref{sec:eos_decision} and {\tab} \ref{tab:scan-results}.

\subsection{CFQ}
On the \textbf{output length split} of CFQ, our Universal Transformer with absolute positional embedding achieves significantly better performance than the one reported in \citet{keysers2020measuring}: $77\%$ versus $\sim 66\%$\footnote{As \citet{keysers2020measuring} only report charts, the exact value is unknown.}.
Here, we were unable to identify the exact reason for this large improvement.
The only architectural difference between the models is
that ours does not make use of any timestep (i.e.~layer ID) embedding.
Also, the positional embedding is only injected to the first layer in case of absolute positional embeddings (Sec.~\ref{sec:model_architectures}).
The relative positional embedding variant performs even better, achieving $81\%$.
This confirms the importance of using relative positional embedding
as a default choice for length generalization tasks, as
we also demonstrated on SCAN in Sec.~\ref{sec:eos_decision}.

On the \textbf{MCD splits}, our results slightly outperform the baseline in \citet{keysers2020measuring}, as shown in {\tab} \ref{tab:results}.
Relative Universal Transformers perform marginally better than all other variants, except for MCD 1 split, where the standard Transformer wins with a slight margin.
We use hyperparameters from \citet{keysers2020measuring}. We report performance after 35k training training steps.

\subsection{PCFG}

The performance of different models on the PCFG dataset is shown on {\tab} \ref{tab:results}.
First of all, simply by increasing the number of training epochs
from 25, used by \citet{hupkes2019compositionality}, to $\sim$237 (300k steps),
our model achieves $65\%$ on the \textbf{productivity split} compared to the $50\%$ reported in \citet{hupkes2019compositionality} and $87\%$ compared to $72\%$ on the \textbf{systematicity split}.
Furthermore, we found that Universal Transformers with relative positional embeddings further improve performance to a large extent, achieving $85\%$ final performance on the \textbf{productivity} and $96\%$ on the \textbf{systematicity split}. We experienced instabilities while training Transformers with relative positional embeddings on the productivity split; thus, the corresponding numbers are omitted in {\tab} \ref{tab:results} and {\fig} \ref{fig:relative} in the appendix.

\subsection{COGS}\label{sec:cogs}

On COGS, our best model achieves the generalization accuracy of $81\%$ which greatly outperforms the $35\%$ accuracy reported in \citet{kim2020cogs}.
This result obtained by simple tricks is competitive compared to the state-of-the-art performance of $83\%$ reported by \citet{akyurek2021lexicon}\footnote{
\citet{akyurek2021lexicon} was published on arXiv on June 7 and later at ACL 2021.
We were not aware of this work at the time of submission to EMNLP 2021 (May 17, 2021).}.
As we discussed in Sec.~\ref{sec:early_stopping}, just by removing early stopping in the setting of \citet{kim2020cogs}, the performance improves to $65\%$. Moreover, the baseline with early stopping is very sensitive to the random seed and even sensitive to the GPU type it is run on. Changing the seed in the official repository from 1 to 2 causes a dramatic performance drop with a $2.5\%$ final accuracy.
By changing the scaling of embeddings (Sec.~\ref{sec:initialization}), disabling label smoothing, fixing the learning rate to $10^{-4}$, we achieved $81\%$ generalization accuracy, which is stable over multiple random seeds.

{\tab} \ref{tab:results} compares different model variants.
Standard Transformers with absolute and relative positional encoding
perform similarly, with the relative positional variant having a slight advantage.
Here Universal Transformers perform slightly worse.

\subsection{Mathematics Dataset}
We also test our approaches on subsets of Mathematics Dataset \cite{saxton2018analysing}.
Since training models on the whole dataset is too resource-demanding,
we only conduct experiments on two subsets: ``place\_value'' and ``add\_or\_sub''.

The results are shown in {\tab} \ref{tab:results}.
While we can not directly compare our numbers with those
reported in \citet{saxton2018analysing}
(a single model is jointly trained on the whole dataset there),
our results show that relative positional embedding is advantageous for the generalization ability on both subsets.

\section{Related Work}
\label{sec:related}
Many recent papers focus on improving generalization on the SCAN dataset. Some of them develop specialized architectures \cite{korrel2019transcoding, li2019compositional, russin2019compositional, gordon2020permutation, herzig2020span} or data augmentation methods \cite{andreas2020good}, others apply meta-learning \cite{lake2019compositional}.
As an alternative, the CFQ dataset proposed in \cite{keysers2020measuring} is gaining attention recently \citep{guo2020hierarchical, furrer2020compositional}.
Mathematical problem solving has also become a popular domain for testing
generalization of neural networks \citep{kaiser2015neural,schlag2019enhancing, charton2021learning}.
The PCFG \citep{hupkes2019compositionality} and COGS \citep{kim2020cogs} are also datasets proposed relatively recently.
Despite increasing interests in systematic generalization tasks, 
interestingly, no prior work has questioned the baseline configurations which could be
overfitted to the machine translation tasks.

Generalizing to longer sequences have been proven to be especially difficult.
Currently only hybrid task-specific neuro-symbolic approaches can solve it \cite{nye2020learning,chen2020compositional,liu2020compositional}.
In this work, we focus on a subproblem required for length generalization: the EOS decision problem \cite{newman2020eos}, and we show that it can be mitigated by using relative positional embeddings.

The study of generalization ability of neural networks at different stages of
training has been a general topic of interest \citep{nakkiran2019deep, roelofs2019measuring}.
Our analysis has shown that this question is particularly relevant to the problem of systematic generalization, as demonstrated by large performance gaps in our experiments,
which has not been discussed in prior work.

Prior work proposed several sophisticated initialization methods for Transformers
\citep{zhang2019improving, zhu2021gradinit},
e.g.~with a purpose of removing the layer normalization components \citep{huang2020improving}.
While our work only revisited basic scaling methods,
we demonstrated their particular importance for systematic generalization.

In recent work,\footnote{Our work was submitted to EMNLP 2021 on May 17, 2021 and has been under the anonymity period until Aug.~25. \citet{ontanon2021making} appeared on arXiv on Aug.~9, 2021.} \citet{ontanon2021making} have also focused on improving the compositional generalization abilities of Transformers.
In addition to relative positional encodings and Universal Transformers, novel architectural changes such as "copy decoder" as well as dataset-specific "intermediate representations" \citep{herzig2021unlocking} have been studied.
However, other aspects we found crucial, such as early stopping, scaling of the positional embeddings, and the validation set issues have not been considered.
In consequence, our models achieve substantially higher performance
than the best results reported by \citet{ontanon2021making} across all standard datasets: PCFG, COGS, and CFQ (without intermediate representations).

Finally, our study focused on the basic Transformer architectures.
However, the \textit{details} discussed above in the context of
algorithmic tasks should also be relevant for
other Transformer variants and fast weight programmers \citep{schmidhuber1992learning, schlag2021linear, irie2021going}, as well as other architectures
specifically designed for algorithmic reasoning \citep{graves2016hybrid, kaiser2015neural, csordas2019improving, freivalds2019neural}. 

\section{Conclusion}
In this work we showed that the performance of Transformer architectures 
on many recently proposed datasets for systematic generalization
can be greatly improved by revisiting basic model and training configurations.
Model variants with relative positional embedding often outperform the ones with absolute positional embedding.
They also mitigate the EOS decision problem, an important problem
previously found by \citet{newman2020eos} when considering the length generalization
of neural networks.
This allows us to focus on the problem of compositions in the future,
which is the remaining problem for the length generalization.

We also demonstrated that reconsidering early stopping and embedding scaling can greatly improve baseline Transformers, in particular on the COGS and PCFG datasets.
These results shed light on the discrepancy between the model performance on the IID validation set and the test accuracy on the systematically different generalization split.
As consequence, currently common practice of validating models on the IID dataset is problematic.
We conclude that the community should discuss proper ways to develop models for systematic generalization.
In particular, we hope that our work clearly demonstrated the necessity of a validation set for systematic generalization in order to establish strong baselines and to avoid a false sense of progress.

\section*{Acknowledgments}

We thank Aleksandar Stanić and Imanol Schlag for their helpful comments and suggestions on an earlier version of the manuscript.
This research was partially funded by ERC Advanced grant no: 742870, project AlgoRNN,
and by Swiss National Science Foundation grant no: 200021\_192356, project NEUSYM.
We thank hardware donations from NVIDIA \& IBM.

\bibliography{anthology,custom}
\bibliographystyle{acl_natbib}

\clearpage
\appendix

\section{Evaluation Metrics}

For all tasks, accuracy is computed on the sequence-level, i.e.~all tokens in the sequence should be correct for the output to be counted as correct.
For the losses, we always report the average token-wise cross entropy loss.

\section{Hyperparameters}

\begin{table*}[ht]
    \centering
    \small
    \caption{Hyperparameters used for different tasks. We denote the feedforward size as $d_\text{FF}$. For the learning rate of CFQ (denoted by *), the learning rate seemingly differs from \citet{keysers2020measuring}.
    In fact, although \citet{keysers2020measuring} use Noam learning rate scheduling, scaling by $\frac{1}{\sqrt{d_\text{model}}}$ is not used, so we had to compensate for this to make them functionally equivalent.}
    \label{tab:hyperparams}
    \begin{tabular}{l c c c c c c c c c}
        \toprule
  & $d_\text{model}$ & $d_\text{FF}$ & $n_\text{head}$ & $n_\text{layers} $ & batch size & learning rate & warmup & scheduler \\
\midrule
SCAN & 128 & 256 & 8 & 3 & 256 & $10^{-3}$ & - & - \\
CFQ - Non-universal & 128 & 256 & 16 & 2 & 4096 & 0.9* & 4000 & Noam \\
CFQ - Universal & 256 & 512 & 4 & 6 & 2048 & 2.24* & 8000 & Noam \\
PCFG & 512 & 2048 & 8 & 6 & 64 & $10^{-4}$ & - & - \\
COGS & 512 & 512 & 4 & 2 & 128 & $10^{-4}$ & - & - \\
COGS Noam & 512 & 512 & 4 & 2 & 128 & 2 & 4000 & Noam \\
Mathematics & 512 & 2048 & 8 & 6 & 256 & $10^{-4}$ & - & - \\
        \bottomrule
    \end{tabular}
\end{table*}

For all of our models we use an Adam optimizer with the default hyperparameters of PyTorch \cite{paszke2019pytorch}. We only change the learning rate. We use dropout with probability of 0.1 after each component of the transformer: both after the attention heads and linear transformations. We specify the dataset-specific hyperparameters in {\tab} \ref{tab:hyperparams}. For all Universal Transformer experiments, we use both the ``No scaling'' and the ``Positional Embedding Downscaling'' methods. For the standard Transformers with absolute positional embedding we test different scaling variants on different datasets shown in {\tab} \ref{tab:init_types}. When multiple scaling methods are available, we choose the best performing ones when reporting results in Table \ref{tab:results}. We always use the same number of layers for both encoder and decoder. The embedding and the final softmax weights of the decoder are always shared (tied embeddings).

The number of parameters for different models and the corresponding to representative execution time is shown in {\tab} \ref{tab:model_size_time}.

\begin{table*}
    \centering
    \small
    \caption{Model sizes and execution times. One representative split is shown per dataset. Other splits have the same number of parameters, and their execution time is in the same order of magnitude.}
    \label{tab:model_size_time}
    \begin{tabular}{l l r r l}
        \toprule
Dataset & Model & No. of params & Execution time & GPU type\\
\midrule
\multirow{4}{*}{SCAN} & Standard & 992k & 1:30 & \multirow{4}{*}{Titan X Maxwell} \\
     & Universal & 333k & 1:15 &  \\
     & Relative Pos. & 1.1M & 1:45 &  \\
     & Universal, Relative Pos. & 366k & 1:30 & \\
\midrule
\multirow{4}{*}{CFQ MCD 2} & Standard & 685k & 10:00 & \multirow{4}{*}{Tesla V100-SXM2-32GB-LS}\\
 & Universal & 1.4M & 12:00 & \\
 & Relative Pos. & 751k & 14:15 & \\
 & Universal, Relative Pos. & 1.5M & 14:00 & \\
\midrule
\multirow{4}{*}{PCFG Systematicity} & Standard & 44.7M & 20:30 & \multirow{4}{*}{Tesla V100-PCIE-16GB}\\
 & Universal & 7.9M & 17:00 & \\
 & Relative Pos. & 47.8M & 21:30 & \\
 & Universal, Relative Pos. & 8.4M & 21:30 & \\
\midrule
\multirow{4}{*}{COGS} & Standard & 9.3M & 17:30 & \multirow{4}{*}{Tesla V100-SXM2-32GB-LS}\\
 & Universal & 5.1M & 17:15 & \\
 & Relative Pos. & 10.3M & 21:00 & \\
 & Universal, Relative Pos. & 5.6M & 20:00 & \\
\midrule
\multirow{4}{*}{Math: add\_or\_sub} & Standard & 4.4M & 8:00 & \multirow{4}{*}{Tesla P100-SXM2-16GB}\\
 & Universal & 7.4M & 7:30 & \\
 & Relative Pos. & 4.7M & 8:30 & \\
 & Universal, Relative Pos. & 7.9M & 8:00 & \\
        \bottomrule
    \end{tabular}
\end{table*}

\begin{table}
    \centering
    \small
    \caption{Scaling types used for standard transformers with absolute positional embedding on different datasets. TEU denotes {\teu}, PED denotes {\ped}.}
    \label{tab:init_types}
    \begin{tabular}{l c c c}
        \toprule
  & TEU & No scaling & PED \\
\midrule
SCAN &   & \checkmark &  \checkmark\\
CFQ MCD &  & \checkmark & \checkmark  \\
CFQ Length & \checkmark & \checkmark  & \checkmark \\
PCFG Productivity & \checkmark & \checkmark & \checkmark \\
PCFG Systematicity & \checkmark & \checkmark  & \checkmark \\
COGS & \checkmark & \checkmark & \checkmark \\
Mathematics &  & \checkmark & \checkmark \\
        \bottomrule
    \end{tabular}
\end{table}

\section{Relative Positional Embedding}\label{sec:rel_pos}

We use the relative positional embedding variant of self attention from \citet{dai2019transformer}. Here, we use a decomposed attention matrix of the following form:

\begin{align*}
\mA_{i, j}^{\mathrm{rel}} &=\underbrace{\mH_i^\top \mW_{q}^\top \mW_{k, E} \mH_j}_{(a)}+\underbrace{\mH_i^\top \mW_{q}^\top \mW_{k, P} \mP_{i-j}}_{(b)} \\
&+\underbrace{\vu^\top \mW_{k, E} \mH_j}_{(c)}+\underbrace{\vv^\top \mW_{k, P} \mP_{i-j}}_{(d)}
\end{align*}
where $\mH_i$ is the hidden state of the $i^\text{th}$ column of the Transformer, $\mP_{i}$ is an embedding for position (or in this case distance) $i$. Matrix $\mW_{q}$ maps the states to queries, $\mW_{k, E}$ maps states to keys, while $\mW_{k, P}$ maps positional embedding to keys. $\vu$ and $\vv$ are learned vectors. Component (a) corresponds to content-based addressing, (b) to content based relative positional addressing, (c) represents a global content bias, while (d) represents a global position bias.

We use sinusoidal positional embedding $\mP_i \in \mathbb{R}^{d_\text{model}}$. The relative position, $i$, can be both positive and negative. Inspired by \citet{vaswani2017attention}, we define $\mP_{i,j}$ as:

\begin{align}
  \mP_{i,j}= 
\begin{cases}
    \sin(i/10000^{2j/d_\text{model}}),& \text{if } j=2k\\
    \cos(i/10000^{2j/d_\text{model}})              & \text{if } j=2k+1
\end{cases}
\label{eq:pos}
\end{align}

Prior to applying the softmax, $\mA_{i, j}^{\mathrm{rel}}$ is scaled by $\frac{1}{\sqrt{d_\text{model}}}$, as in \citet{vaswani2017attention}.

We never combine absolute with relative positional embedding. In case of a relative positional variant of any Transformer model, we do not add absolute positional encoding to the word embeddigs. We use relative positional attention in every layer, except at the interface between encoder and decoder, where we use the standard formulation from \citet{vaswani2017attention}, without adding any positional embedding.

\section{Embedding Scaling}
\label{app:scaling}
In this section, we provide full descriptions of
embedding scaling strategies we investigated.
In the following, $w_i$ denotes the word index at input position $i$, $\mE_w \in \mathbb{R}^{d_\text{model}}$ denotes learned word embedding for word index $w$. Positional embedding for position $i$ is defined as in Eq. \ref{eq:pos}.

\paragraph{\teu.}
\citet{vaswani2017attention} combine the input word and
positional embeddings for each position $i$ as $\mH_i = \sqrt{d_\text{model}} \mE_{w_i} + \mP_i$.
Although in the original paper, the initialization of $\mE$ is not discussed, most implementations use Glorot initialization \cite{glorot2010understanding},
which in this case means that each component of $\mE$ is drawn from $\mathcal{U}(-\sqrt{\frac{6}{d_\text{model} + N_\text{words}}}, \sqrt{\frac{6}{d_\text{model} + N_\text{words}}})$
where $\mathcal{U}(a,b)$ represents the uniform distribution in range $[a,b]$.

\paragraph{No scaling.} This corresponds to how PyTorch initializes embedding layers by default: each element of $\mE$ is drawn from $\mathcal{N}(0,1)$. $\mathcal{N}(\mu, \sigma)$ is the normal distribution with mean $\mu$ and standard deviation of $\sigma$. The word embeddings are combined with the positional embeddings without any scaling: $\mH_i = \mE_{w_i} + \mP_i$

\paragraph{\ped.} We propose to use Kaiming initialization \cite{he2015delving} for the word embeddings:~each element of $\mE \sim \mathcal{N}(0,\frac{1}{\sqrt{d_\text{model}}})$. Instead of scaling up the word embeddings, the positional embeddings are scaled down: $\mH_i = \mE_{w_i} + \frac{1}{\sqrt{d_\text{model}}}\mP_i$

\begin{table*}
\setlength{\tabcolsep}{0.47em}
    \centering
    \small
    \caption{Test accuracy of different Transformer (Trafo) variants and different initializations on the considered datasets. This is a more detailed version of Table \ref{tab:results}, with detailed scores for all initialization variants. The last column shows previously reported accuracies. References: [1] \citet{newman2020eos}, [2] \citet{keysers2020measuring},  [3] \url{https://github.com/google-research/google-research/tree/master/cfq}, [4] \citet{hupkes2019compositionality}, [5] \citet{kim2020cogs}, [6] \citet{saxton2018analysing}. Results marked with $*$ cannot be directly compared because of different training setups. $\sim$ denotes imprecise numbers read from charts in prior works. For the configuration marked by $\dagger$, the results are obtained by running 8 seeds from which 3 crashed, resulting in 5 useful runs reported below. Crashed runs suddenly drop their accuracy to 0, which never recovers during the training. The reason for the crashing is the overly big learning rate (2.24, from the baseline). We run another 10 seeds with learning rate of 2.0, obtaining similar final accuracy of $0.75 \pm 0.02$, but without any crashed runs.}
    \label{tab:results_init_details}
    \begin{tabular}{l l c c c c | c}
        \toprule
  & Init & Trafo & Uni. Trafo & Rel. Trafo & Rel. Uni. Trafo & Reported\\
\midrule
\multirow{2}{*}{SCAN (length cutoff=26)} & PED & \bf{0.30 $\pm$ 0.02} & \bf{0.21 $\pm$ 0.01} & - & - & \multirow{2}{*}{$0.00^{[1]}$} \\ 
 & No scaling & 0.15 $\pm$ 0.07 & 0.14 $\pm$ 0.05 & \bf{0.72 $\pm$ 0.21} & \bf{1.00 $\pm$ 0.00} & \\ 
\midrule
\multirow{3}{*}{CFQ Output length} & PED & 0.56 $\pm$ 0.02 & 0.60 $\pm$ 0.34 & - & - & \multirow{3}{*}{$\sim 0.66^{[2]}$} \\ 
 & TEU & \bf{0.57 $\pm$ 0.00} & \hspace{1.35mm} 0.74 $\pm$ 0.02 $\dagger$ & - & - & \\ 
 & No scaling & 0.53 $\pm$ 0.04 & \bf{0.77 $\pm$ 0.02} & \bf{0.64 $\pm$ 0.06} & \bf{0.81 $\pm$ 0.01} & \\ 
\cmidrule{2-7}
\multirow{2}{*}{CFQ MCD 1} & PED & 0.36 $\pm$ 0.02 & 0.37 $\pm$ 0.05 & - & - & \multirow{2}{*}{$0.37\pm0.02^{[3]}$} \\ 
 & No scaling & \bf{0.40 $\pm$ 0.01} & \bf{0.39 $\pm$ 0.03} & \bf{0.39 $\pm$ 0.01} & \bf{0.39 $\pm$ 0.04} & \\ 
\cmidrule{2-7}
\multirow{2}{*}{CFQ MCD 2} & PED & 0.08 $\pm$ 0.01 & \bf{0.09 $\pm$ 0.01} & - & - & \multirow{2}{*}{$0.08\pm0.02^{[3]}$} \\ 
 & No scaling & \bf{0.10 $\pm$ 0.01} & \bf{0.09 $\pm$ 0.02} & \bf{0.09 $\pm$ 0.01} & \bf{0.10 $\pm$ 0.02} & \\ 
\cmidrule{2-7}
\multirow{2}{*}{CFQ MCD 3} & PED & 0.10 $\pm$ 0.00 & \bf{0.11 $\pm$ 0.00} & - & - & \multirow{2}{*}{$0.11\pm0.00^{[3]}$} \\ 
 & No scaling & \bf{0.11 $\pm$ 0.00} & \bf{0.11 $\pm$ 0.01} & \bf{0.11 $\pm$ 0.01} & \bf{0.11 $\pm$ 0.03} & \\ 
\cmidrule{2-7}
\multirow{2}{*}{CFQ MCD mean} & PED & 0.18 $\pm$ 0.13 & 0.19 $\pm$ 0.14 & - & - & \multirow{2}{*}{$0.19\pm0.01^{[2]}$} \\ 
 & No scaling & \bf{0.20 $\pm$ 0.14} & \bf{0.20 $\pm$ 0.14} & \bf{0.20 $\pm$ 0.14} & \bf{0.20 $\pm$ 0.14} & \\ 
\midrule
\multirow{3}{*}{PCFG Productivity split} & PED & \bf{0.65 $\pm$ 0.03} & \bf{0.78 $\pm$ 0.01} & - & - & \multirow{3}{*}{$0.50\pm0.02^{[4]}$} \\ 
     & TEU & 0.47 $\pm$ 0.27 & \bf{0.78 $\pm$ 0.01} & - & - & \\ 
 & No scaling & 0.63 $\pm$ 0.02 & 0.76 $\pm$ 0.01 & - & \bf{0.85 $\pm$ 0.01} & \\ 
\cmidrule{2-7}
\multirow{3}{*}{PCFG Systematicity split} & PED & \bf{0.87 $\pm$ 0.01} & \bf{0.93 $\pm$ 0.01} & - & - & \multirow{3}{*}{$0.72\pm0.00^{[4]}$} \\ 
 & TEU & 0.75 $\pm$ 0.08 & 0.92 $\pm$ 0.01 & - & - & \\ 
 & No scaling & 0.86 $\pm$ 0.02 & 0.92 $\pm$ 0.00 & \bf{0.89 $\pm$ 0.02} & \bf{0.96 $\pm$ 0.01} & \\ 
\midrule
\multirow{3}{*}{COGS} & PED & \bf{0.80 $\pm$ 0.00} & 0.77 $\pm$ 0.02 & - & - & \multirow{3}{*}{$0.35\pm0.06^{[5]}$} \\ 
 & TEU & 0.78 $\pm$ 0.03 & \bf{0.78 $\pm$ 0.03} & - & - & \\ 
 & No scaling & 0.62 $\pm$ 0.06 & 0.51 $\pm$ 0.07 & \bf{0.81 $\pm$ 0.01} & \bf{0.77 $\pm$ 0.01} & \\ 
\midrule
\multirow{2}{*}{Math: add\_or\_sub} & PED & 0.80 $\pm$ 0.01 & 0.92 $\pm$ 0.02 & - & - & \multirow{2}{*}{$\sim0.91^{[6]*}$} \\ 
 & No scaling & \bf{0.89 $\pm$ 0.01} & \bf{0.94 $\pm$ 0.01} & \bf{0.91 $\pm$ 0.03} & \bf{0.97 $\pm$ 0.01} & \\ 
\cmidrule{2-7}
\multirow{2}{*}{Math: place\_value} & PED & 0.00 $\pm$ 0.00 & \bf{0.20 $\pm$ 0.02} & - & - & \multirow{2}{*}{$\sim0.69^{[6]*}$} \\ 
 & No scaling & \bf{0.12 $\pm$ 0.07} & 0.12 $\pm$ 0.01 & - & \bf{0.75 $\pm$ 0.10} & \\ 
        \bottomrule
    \end{tabular}
\end{table*}

\section{Analyzing the Positively Correlated Loss and Accuracy}

\begin{figure}
\begin{center} 
\subfloat[Decomposed loss]{\includegraphics[width=0.95\columnwidth]{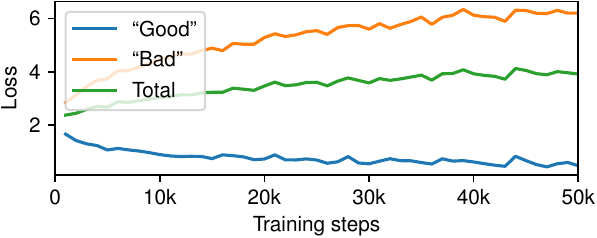}}\\
\subfloat[Histogram of ``good'' loss (first and last measurement)]{\includegraphics[width=0.95\columnwidth]{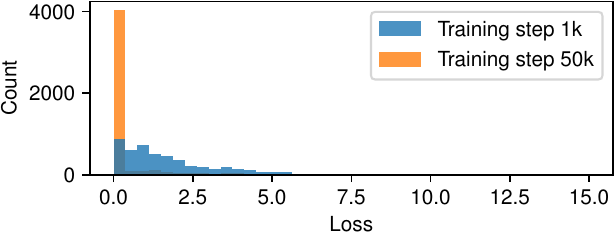}}\\
\subfloat[Histogram of ``bad'' loss (first and last measurement)]{\includegraphics[width=0.95\columnwidth]{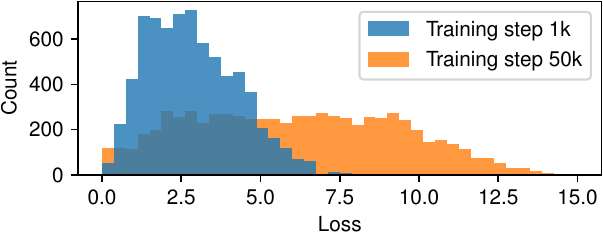}}

\caption{Analysis of the growing test loss on the systematically different test set on CFQ MCD 1 split. We measure the loss individually for each sample in the test set. We categorize samples as ``good'' if the network output on the corresponding input matched the target exactly any point during the training, and as ``bad'' otherwise. (a) The total loss (increasing) can be decomposed to the loss of the ``good'' samples (decreasing), and the loss of the ``bad'' samples (increasing). (b, c) The histogram of the loss for the ``good'' and ``bad'' samples at the beginning and end of the training. The loss of the ``good'' samples concentrates near zero, while the ``bad'' samples spread out and the corresponding loss can be very high. The net effect is a growing total loss.
}\label{fig:loss_grow_analysis}

\end{center}
\end{figure}

In Sec.~\ref{sec:loss_and_accuracy}, we reported 
that on the generalization splits of some datasets both the accuracy and the loss grows together during training.
Here we further analyze this behavior in {\fig} \ref{fig:loss_grow_analysis} (see the caption).

\begin{figure}
\begin{center}
\subfloat[COGS: IID Validation set]{\includegraphics[width=0.95\columnwidth]{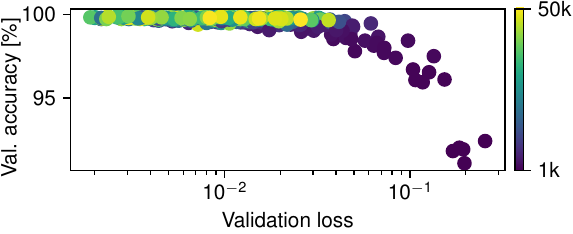}}
\\
\subfloat[COGS: Generalization test set]{\includegraphics[width=0.95\columnwidth]{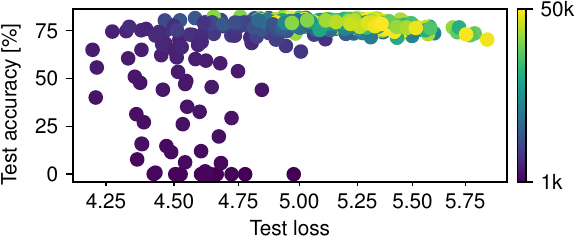}}

\caption{Relationship between the loss and accuracy on (a) IID validation set and (b) the generalization test set on COGS (it comes without a validation set for the generalization splits). Standard Transformers are used. The color shows the training step. Five runs are shown. The loss is shown on a logarithmic scale. On the IID validation set (a), the accuracy increases when the loss decreases, as expected. In contrast, on the generalization split (b), high accuracy corresponds to higher loss. For generalization validation loss versus generalization accuracy on CFQ MCD 1, see {\fig} \ref{fig:loss_grows}. For the analysis of the underlying reason, see {\fig} \ref{fig:loss_grow_analysis}.
}\label{fig:loss_grows_details}

\end{center}
\end{figure}

\section{Accuracies on the IID Split}\label{sec:iid_accuracy}

To show that the IID accuracy does not provide any useful signal for assessing the quality of the final model, we report IID accuracies of the models from {\tab} \ref{tab:results} in {\tab} \ref{tab:results_iid}.
We only show datasets for which an IID validation set is available in the same split as the one reported in {\tab} \ref{tab:results}.
This complements the IID and generalization accuracies on COGS and PCFG with different embedding scalings we reported in {\tab} \ref{tab:init_iid}.
With the exception of standard Transformer on PCFG and the ``place\_value'' module of the Mathematics dataset, all other validation accuracies are 100\%, while their generalization accuracy vary wildly.

\begin{table*}
    \centering
    \small
    \caption{IID validation accuracy for datasets where IID test set is available. CFQ and PCFG are not shown because they require the model to be trained on a separate, IID split. The other settings correspond to {\tab} \ref{tab:results} in the main text. Generalization split test accuracies are shown in parenthesis for easy comparison.}
    \label{tab:results_iid}
    \begin{tabular}{l c c c c }
        \toprule
 & Transformer & Uni. Transformer & Rel. Transformer & Rel. Uni. Transformer \\
\midrule
SCAN (length cutoff=26) & {\bf1.00 $\pm$ 0.00} (0.30) & {\bf1.00 $\pm$ 0.00} (0.21) & {\bf1.00 $\pm$ 0.00} (0.72) & {\bf1.00 $\pm$ 0.00} (1.00) \\
\midrule
COGS & {\bf1.00 $\pm$ 0.00} (0.80) & {\bf1.00 $\pm$ 0.00} (0.78) & {\bf1.00 $\pm$ 0.00} (0.81) & {\bf1.00 $\pm$ 0.00} (0.77) \\
\midrule
Math: add\_or\_sub & {\bf1.00 $\pm$ 0.00} (0.89) & {\bf1.00 $\pm$ 0.00} (0.94) & {\bf1.00 $\pm$ 0.00} (0.91) & {\bf1.00 $\pm$ 0.00} (0.97) \\
Math: place\_value & 0.80 $\pm$ 0.45 (0.12) & {\bf1.00 $\pm$ 0.00} (0.20) & - & {\bf1.00 $\pm$ 0.00} (0.75) \\
        \bottomrule
    \end{tabular}
\end{table*}

\section{Additional Results}

{\fig} \ref{fig:loss_grows_details} shows that both the test loss and accuracy grows on COGS dataset during training. Additionally, it shows the expected, IID behavior on the same dataset for contrast.

\begin{figure}
\begin{center}
\includegraphics[width=0.95\columnwidth]{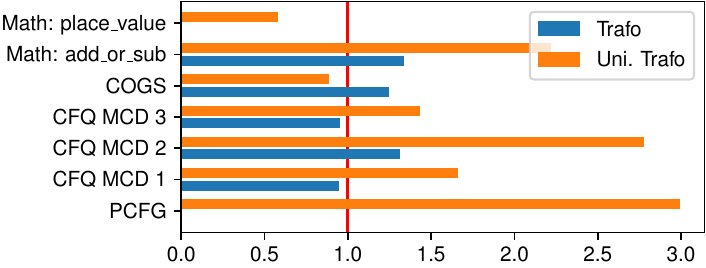}\label{fig:relspeed}
\caption{Relative change in convergence speed by using relative positional embeddings instead of absolute. Convergence speed is measured as the mean number of steps needed to achieve $80\%$ of the final performance of the model. Relative variants usually converge faster. Universal Transformers benefit more than the non-universal ones. The non-universal variants are not shown for PCFG and ``Math: place\_value'', because the relative variants do not converge (see Sec.~\ref{sec:further_benefits}).
}\label{fig:relative}
\end{center}
\end{figure}

{\fig} \ref{fig:relative} shows the relative change in convergence speed when using relative positional embeddings.

\begin{table*}
    \centering
    \small
    \caption{Accuracy of different Transformer variants on CFQ. ``Big'' variant has a batch size of 4096, and is trained with Noam scheduler (learning rate 0.9). ``Small'' variant has a batch size of 512 and a fixed learning rate of $10^{-4}$. The ratio of accuracies of ``small'' and ``big'' variants are also shown in the ``Ratio'' column, indicating the relative performance drop caused by decreasing the batch size. Relative variants experience less accuracy drop.}
    \begin{tabular}{l l c c c c}
        \toprule
& Variant & Transformer & Rel. Transformer & Uni. Transformer & Rel. Uni. Transformer\\
\midrule
\multirow{3}{*}{CFQ MCD 1} & Big & $0.40\pm0.01$ & $0.39\pm0.02$ & $0.41\pm0.03$ & $0.42\pm0.02$\\
 & Small & $0.26\pm0.02$ & $0.32\pm0.01$ & $0.28\pm0.00$ & $0.36\pm0.01$\\
\cmidrule{2-6}
 & Ratio & 0.65 & 0.80 & 0.68 & \bf0.85\\
\midrule
\multirow{3}{*}{CFQ MCD 2} & Big & $0.10\pm0.01$ & $0.09\pm0.01$ & $0.09\pm0.00$ & $0.09\pm0.02$\\
 & Small & $0.05\pm0.01$ & $0.07\pm0.01$ & $0.04\pm0.01$ & $0.10\pm0.01$\\
\cmidrule{2-6}
 & Ratio & 0.51 & 0.76 & 0.50 & \bf1.05\\
\midrule
\multirow{3}{*}{CFQ MCD 3} & Big & $0.11\pm0.00$ & $0.11\pm0.01$ & $0.11\pm0.01$ & $0.12\pm0.02$\\
 & Small & $0.09\pm0.00$ & $0.09\pm0.00$ & $0.09\pm0.01$ & $0.11\pm0.01$\\
\cmidrule{2-6}
 & Ratio & 0.80 & 0.85 & 0.85 & \bf0.98\\
\midrule
\multirow{3}{*}{CFQ Out. len.} & Big & $0.57\pm0.02$ & $0.64\pm0.04$ & $0.76\pm0.03$ & $0.81\pm0.02$\\
 & Small & $0.41\pm0.03$ & $0.51\pm0.02$ & $0.55\pm0.02$ & $0.70\pm0.03$\\
\cmidrule{2-6}
 & Ratio & 0.72 & 0.80 & 0.73 & \bf0.87\\
        \bottomrule
    \end{tabular}
    \label{tab:small_batch_details_abs}
\end{table*}

\begin{table*}
    \centering
    \small
    \caption{Dataset statistics. ``\#'' denotes number of samples. Vocabulary size shows the union of input and output vocabularies. Train and test length denotes the maximum input/output length in the train and test set, respectively.}
    \label{tab:dataset_stat}
    \begin{tabular}{l c c c c c c c}
        \toprule
Dataset & \# train & \# IID valid. & \# gen. test & \# gen. valid. & Voc. size & Train len. & Test len.\\
\midrule
Scan (length cutoff=26)  & $16458$ & $1828$ & $2624$ & - & $19$ & $9$/$26$ & $9$/$48$ \\
\midrule
CFQ MCD 1  & $95743$ & - & $11968$ & $11968$ & $181$ & $29$/$95$ & $30$/$103$ \\
CFQ MCD 2  & $95743$ & - & $11968$ & $11968$ & $181$ & $29$/$107$ & $30$/$91$ \\
CFQ MCD 3  & $95743$ & - & $11968$ & $11968$ & $181$ & $29$/$107$ & $30$/$103$ \\
CFQ Output Length  & $100654$ & - & $9512$ & $9512$ & $181$ & $29$/$77$ & $29$/$107$ \\
\midrule
PCFG Productivity  & $81010$ & - & $11333$ & - & $535$ & $53$/$200$ & $71$/$736$ \\
PCFG Systematicity  & $82168$ & - & $10175$ & - & $535$ & $71$/$736$ & $71$/$496$ \\
\midrule
COGS  & $24155$ & $3000$ & $21000$ & - & $871$ & $22$/$153$ & $61$/$480$ \\
\midrule
Math: add\_or\_sub  & $1969029$ & $10000$ & $10000$ & - & $69$ & $60$/$19$ & $62$/$23$ \\
Math: place\_value  & $1492268$ & $9988$ & $10000$ & - & $69$ & $50$/$1$ & $52$/$1$ \\
        \bottomrule
    \end{tabular}
\end{table*}

\end{document}

%% file: math_commands.tex
\usepackage{amsmath,amsfonts,bm}

{}
{}
{}
{}

\def\eqref#1{equation~\ref{#1}}

\def\1{\bm{1}}

\def\vu{{\bm{u}}}
\def\vv{{\bm{v}}}

\def\mA{{\bm{A}}}

\def\mE{{\bm{E}}}

\def\mH{{\bm{H}}}

\def\mP{{\bm{P}}}

\def\mW{{\bm{W}}}

\DeclareMathAlphabet{\mathsfit}{\encodingdefault}{\sfdefault}{m}{sl}
\SetMathAlphabet{\mathsfit}{bold}{\encodingdefault}{\sfdefault}{bx}{n}

